\documentclass[11pt,a4paper]{article}
\usepackage[hyperref]{acl2019}
\usepackage{times}
\usepackage{latexsym}
\usepackage{url}
\usepackage{amssymb,amsmath}
\usepackage{graphicx,float}
\usepackage{paralist}

\aclfinalcopy

\title{Low-resource Languages:\\A Review of Past Work and Future Challenges}

\author{
	Alexandre Magueresse, Vincent Carles, Evan Heetderks \\
	Department of Computer Science and Technology \\
	Tsinghua University, Beijing, China \\
	\texttt{\{maihr19, wenst19, heetderksej11\}@mails.tsinghua.edu.cn} \\
}

\date{\today}

\begin{document}
\maketitle

\begin{abstract}
A current problem in NLP is massaging and processing low-resource languages which lack useful training attributes such as supervised data, number of native speakers or experts, etc. This review paper concisely summarizes previous groundbreaking achievements made towards resolving this problem, and analyzes potential improvements in the context of the overall future research direction.
\end{abstract}

\section{Introduction}

In the 1990s, the tools of Natural Language Processing (NLP) experienced a major shift, transitioning from rule-based approaches to statistical-based techniques. Most of today's NLP research focuses on 20 of the 7000 languages of the world, leaving the vast majority of languages understudied. These languages are often referred to as low-resource languages (LRLs), ill-defined as this qualifier can be.

LRLs can be understood as \emph{less studied}, \emph{resource scarce}, \emph{less computerized}, \emph{less privileged}, \emph{less commonly taught}, or \emph{low density}, among other denominations \cite{singh2008natural,cieri2016selection,tsvektov2017opportunities}. In this paper, the term LRL will refer to languages for which statistical methods cannot be directly applied because of data scarcity.

There are many reasons to care about LRLs. Africa and India are the hosts of around 2000 LRLs, and are home to more than 2.5 billion inhabitants. Developing technologies for these languages opens considerable economic perspectives. Also, supporting a language with NLP tools can prevent its extinction and foster its expansion, open the knowledge contained in original works to everyone, or even expand prevention in the context of emergency response. \cite{tsvektov2017opportunities}

This paper provides an overview of the recent methods that have been applied to LRLs as well as underlines promising future direction and points out unsolved questions.

\section{Related work}

To the best of our knowledge, previous work that conducted reviews on LRLs either aimed attention at specific tasks such as parts-of-speech tagging \cite{christodoulopoulos2010two}, text classification \cite{cruz2020establishing} and machine translation \cite{lakew2020low}, or, as the illustrious work on textual analysis (including named entities, parts-of-speech, morphological analysis) \cite{yarowsky2001inducing} can tell, trace back to decades ago.

The promotion of LRLs has also been at the core of several conferences and workshops such as LREC \footnote{\url{www.lrec-conf.org/}}, AMTA \footnote{\url{www.amtaweb.org/}} or LoResMT \footnote{\url{https://www.mtsummit2019.com/}}.

\section{The projection technique}

We start by presenting the projection technique, that is central to most NLP tasks applied to LRLs. The rest of our paper examines resource collection with a focus on automatic alignment (section 4), linguistic tasks, which directly take on the projection technique (section 5), speech recognition (section 6), multilingual embeddings (section 7) and their application to machine translation (section 8) and classification (section 9). We conclude by discussing the need for a unified framework to assess the linguistic closeness of two languages, as well as evaluate the ability of a solution to generalize over LRLs, and call further studies to work on more diversified languages.

The projection, or alignment technique has been massively used by the literature since its formalization by \cite{yarowsky2001inducing}. Indeed, it enables to make the most of the existing annotations of a high-resource language (HRL) and apply it on a LRL, for which annotation is either hard to collect or simply impossible for lack of language expert. 

\begin{figure}
    \centering
    \includegraphics[width=\linewidth]{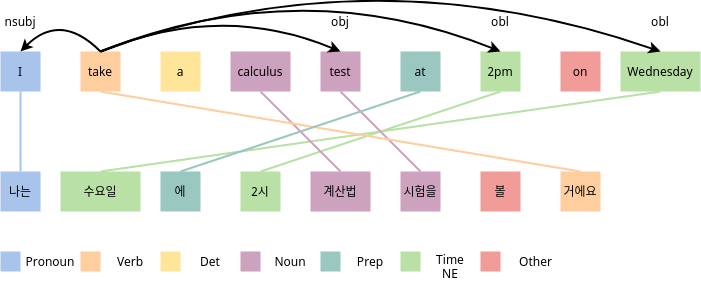}
    \caption{Sample projection from English to Korean}
    \label{fig:projection}
\end{figure}

There are three levels of alignment, document, sentence and word, and each brings about increasing information (in that order) at the cost of more complex annotation. To list a few examples, document-level alignment is useful in information retrieval, sentence-level alignment is at the core of machine translation (section 8), and word-level alignment grounds most linguistic tasks (section 5). Figure \ref{fig:projection} provides a sample word-level alignment between English and Korean.

It is difficult to collect corpora aligned at the word level. This is why growing efforts are pursued at automatically aligning corpora, with a focus on word alignment. We further discuss such techniques in section 4.2.

Typically, the ideal one-to-one mapping setting is the exception rather than the rule: there are many cases where the source and target languages exhibit mismatching structures, as can be seen on Figure \ref{fig:projection}. There are indeed two major prerequisites to ensure annotations can be efficiently projected:
\begin{inparaenum}[i)]
    \item the two involved languages should be structurally (grammatically) close to each other depending on the task (for example both languages should share the same inventory of tag sets for part-of-speech tagging)
    \item the underlying annotations of both languages have to be consistent with each other.
\end{inparaenum}

\section{Resource collection}

By definition, LRLs lack resources that are still necessary for any NLP task. There are two general trends to collecting information for LRLs: 
\begin{inparaenum}[i)]
    \item creating new datasets by annotating raw text
    \item gathering raw text and aligning it with a higher-resource language
\end{inparaenum}

\subsection{Dataset creation}

A constant line of efforts has addressed the question of dataset creation, for languages from all around the globe.

First, the REFLEX-LCTL\footnote{Research on English and Foreign Language Exploitation -- Less Commonly Taught Languages} \cite{simpson2008human} and LORELEI\footnote{LOw-REsource Languages for Emergent Incidents}\cite{strassel2016lorelei} projects, conducted by the Linguistic Data Consortium (LDC), released annotated corpora for 13 and 34 languages respectively. LORELEI went a step further by integrating each dataset in a unified framework that includes entity, semantic, part-of-speech and noun phrase annotation.

The challenges faced by both LDC projects  served as a lesson for many other programs. \cite{godard2017very} produced a speech corpus that can be used for speech recognition and language documentation. \cite{marivate2020investigating,marivate2020low} curated a collection of news headlines for two South African languages, as well as established baselines for the classification task. This work showed what was possible to achieve using very limited amounts of data and should be an example to guide future work in collecting new dataset for LRLs.

\paragraph{Future work} The aforementioned studies proved greatly innovative in the sources they examined to built their datasets. Social media, mobile applications but also governmental documents can all be taken advantage of to generate textual content.

\subsection{Automatic alignment}

In the typical scenario, automatic alignment requires language expertise and is time-consuming. However, since most techniques involving LRLs rely on aligned corpora, it has been the research focus of many recent studies.

\begin{figure}
    \centering
    \includegraphics[width=0.8\linewidth]{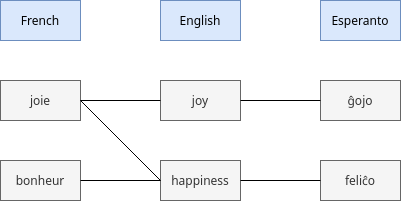}
    \caption{Ambiguity problem of the pivot technique}
    \label{fig:lexicon}
\end{figure}

\paragraph{Word-level alignment} The most elementary alignment scope is at the word-level ; this task is most commonly referred to as lexicon induction. The inverse consultation (IC) method \cite{saralegi2011analyzing}, that builds and compares both dictionaries $A \to C$ and $C \to A$ from dictionaries $A \leftrightarrow B$ and $B \leftrightarrow C$, has been used in varied contexts, but fails to model lexical variants and polysemy (Figure \ref{fig:lexicon}, ``joie'' will be translated as ``feliĉo'' whereas it should not). The distributional similarity (DS) method examines the context of a word to further improve the IC algorithm.

The more recent cognate equivalence assumption, that claims that two words sharing similar writing, meaning and etymology share all their meanings, significantly helped develop lexicon and synonym dictionaries \cite{nasution2016constraint,nasution2017generalized}. Given two dictionaries $A \to B$ and $B \to C$, other work introduced constraints on the similarity of languages $A$ and $C$ and modeled the structures of the input dictionaries as a Boolean optimization problem, in order to build dictionary $A \to C$ \cite{wushouer2015constraint}.

\paragraph{Sentence-level alignment} Particularly useful for machine translation, sentence-level alignment has recently gained momentum. The general method consists in two steps: devising a similarity score between two sentences and aligning sentences based on these scores. A great variety of methods aimed to evaluate the distance between two sentences. Since the release of the WMT16 dataset \footnote{\url{http://www.statmt.org/wmt16/}}, translating the source sentence into the target language and matching n-grams \cite{dara2016yoda,gomes2016first,shchukin2016word}, or using cosine similarity on tf-idf vectors \cite{buck2016quick,jakubina2016bad} has become popular. Even if word-to-word translations can be applied to LRLs, these systems often require higher-quality translation. Consequently, another trend preferably relies on multilingual embedding such as \cite{artetxe2019massively} and derive a similarity score with the word mover distance (WMD) \cite{huang2016supervised,mueller2019sentence}.

\paragraph{Document-level alignment} Aligning corpora at the document-level is useful for text classification, translation or multilingual representations. Based on sentence embedding and more relaxed formulations of the Earth mover's distance, \cite{el2020massively} obtained state-of-the-art alignment on low- and mid-resource languages.

\paragraph{Future work} More emphasis should be put on sentence distance evaluation, since it directly influences current results for document alignment. Other solutions to improve document alignment could include designing new weighting schemes to account for significant differences in document sizes. Finally, heuristics could help cut down the complexity of document alignment methods, such as relative position in the document or sentence length.

\section{Linguistic tasks}

The ``linguistic'' tasks we consider in this paper are mostly related to grammar modeling, and their applications serve the interest of linguistics and research rather than any commercial purpose (such as machine translation, speech recognition, summarization and many other).

\subsection{Part-of-speech tagging}

Solving the part of speech (POS) tagging task in an unsupervised fashion essentially amounts to a clustering task. Once every word is assigned to one or several clusters, each cluster needs to be grounded, or mapped to its corresponding POS. Grounding fatally requires a minimum annotation.

There are many clustering algorithms, of which Brown \cite{brown1992class} is one of the earliest formulation. It has yet been shown that this simple clustering technique is often the one that leads to best performance \cite{christodoulopoulos2010two}. Another very common technique inherits the hidden Markov model (HMM) and views the POS tagging task as a sequence-to-sequence problem. By training on a ``parent'' language for which annotations are available, it is even possible to operate the grounding step without annotation for the LRL, provided that the tags are standardized among both languages \cite{buys2016cross,cardenas2019grounded}. Both works showed that the choice of the ``parent'' language is difficult, because typologically close languages do not always work best in practise, even when relying on expert features provided by the WALS \footnote{\url{www.wals.info}} features. Instead, it advocates combining several parent languages to better leverage word-order patterns, and shirk the question of parent language selection.

Supervised (or semi-supervised) methods are also applied to the POS tagging task, using projection to cope with the lack of annotations. Earliest methods restricted their application to closely-related languages by directly tagging the target sentence the same way as the source sentence (also including class rebalancing to account for minor linguistic differences) \cite{yarowsky2001inducing}. Long short-term memory networks (LSTM) based on multilingual embeddings proved efficient when available annotations were in larger quantities, proved \cite{zennaki2015utilisation}.

\begin{figure}
    \centering
    \includegraphics[width=\linewidth]{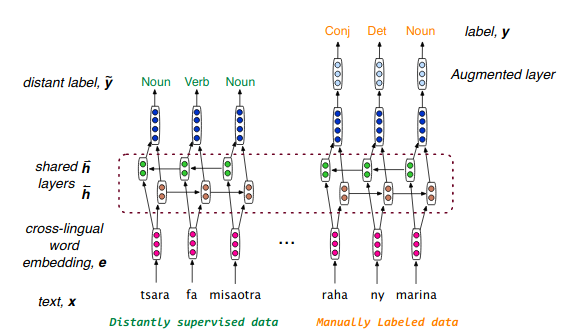}
    \caption{One model architecture for POS-tagging, from \cite{fang2017model}}
    \label{fig:pos}
\end{figure}

POS-tagging as a classification problem was also attacked with hand-crafted features on very limited amounts of annotations, backed with reasonable help of projection on the English language \cite{duong2014can}. The proposed model is a softmax classifier, first trained on projected annotations, and then adjusted on ground-truth tags. Two other formulations relying on bidirectional LSTMs further improved tagging accuracy on some languages \cite{fang2016learning,fang2017model}. As can be seen on Figure \ref{fig:pos}, the output of the BiLSTM model is either directly used for the distantly annotated data, or first passed through a projection layer to be evaluated on the ground truth tags of the LRL. The two formulations differ in the way the hidden states of the BiLSTM are 
projected (either matrix multiplication \cite{fang2016learning} or MLP \cite{fang2017model}).

\paragraph{Future work} Most projected annotations currently stem from English corpora. One direction for future work would be relying one or several other HRLs to transfer annotations from. Also, it might be interesting to find a way to combine the information of multiple LRLs simultaneously, especially if they are closely-related. Besides, a more robust evaluation metric has to be adopted by unsupervised methods for POS tagging. \cite{christodoulopoulos2010two} suggests using the V-measure, analogous to the F-score metrics.

\subsection{Dependency parsing}

This section first summarizes two important findings that greatly influenced the following studies. Then, it shows how parameter sharing and zero-shot transfer are profitable to dependency parsing. Both techniques are facilitated by unified annotation schemes, such as the Universal Dependencies (UD) dataset \footnote{\url{www.universaldependencies.org}}.

\paragraph{Two major findings}
Dependency parsing has traditionally built on POS-tagging, but in the context of LRLs, gold POS tags are not available. Replacing POS information by inferred word clusters is often more beneficial than trying to predict gold POS \cite{spitkovsky2011unsupervised}. Allowing words to belong to different clusters in different contexts indeed helps model polysemy. Another study showed that transferring from multiple sources is generally preferable to single source when languages share grammatical similarities \cite{mcdonald2011multi} ; not because more data is leveraged but because the model is thus exposed to a variety of patterns.

\paragraph{Parameter sharing} There are even some cases where distant supervision through jointly learning dependency parsing on a source and target languages leads to more accurate performance for the target language, than relying on supervised data for the target language \cite{duong2015low,duong2015neural}. These two papers let their model share parameters across two models, and only the embedding layer is language-dependent.

\paragraph{Zero-shot transfer} Transition-based LSTMs predicting sequential manipulations on a sentence (reduce left or right, shift) taking cluster and POS information as well as embedding as input outperforms previous cross-lingual multi-source model transfer methods \cite{ammar2016many}. Following the multi-source transfer guideline, \cite{agic2016multilingual} projects annotations weighted by alignment probability scores. This weighted scheme averaged on a large number of languages reached unprecedented unlabeled attachment score (UAS) for more than 20 languages. \cite{wu2018multilingual} establishes that carefully selecting source languages allows for efficient direct transfer learning from models that perform well on HRLs. This naturally raises the question of how to optimally select proper source languages given a LRL.

\paragraph{Future work} Several studies suggest learning POS and dependencies in a joint fashion would incite major advances in both tasks. \cite{lim2020semi} recently presented a multi-view learning framework that jointly learns POS tagging and parsing. Both token and character embeddings feed two independent LSTMs, the hidden state of which are used as input as another central LSTM. The outputs of the three models are then combined and co-trained on both tasks.

Besides, sharing parameters across closely-related languages could give rise to further studies on larger scales and more diverse language sets. More generally, the question of language ``closeness'' is a topical issue that needs careful consideration.

\subsection{Named entity recognition, typing and linking}

Extracting (recognizing, NER) named entities from a text, classifying them (typing, NET) into categories (such as location, name or organization) and building bridges across them (linking, NEL) is of capital importance in information retrieval, recommendation system and classification. In the context of LRLs, these three tasks have been the subject of increasing research.

\paragraph{NER and NET} Both tasks are often jointly learnt as a classification task which includes a category for wrods that are not named entities. Projection has been widely used in this task, either to learn a model on a HRL and applying it to a LRL \cite{zamin2020projecting}, or to train a classifier on the projected annotations from a HRL \cite{yarowsky2001inducing}. The early solutions to solve this task mainly involved Hidden Markovian Models (HMM), but they fail to handle named entity phrases and cannot model non-local dependencies along the sentence \cite{yarowsky2001inducing}. A transition towards Conditional Random Fields (CRF) helped answer these two issues, at the cost of hand-made and language-dependent features \cite{saha2008hybrid,demir2014improving,littell2016named}. The features not only involved prefixes and suffixes, but also more language-specific morphology induction methods and transcriptions in the international phonetic alphabet.

More recent work replaced costly feature engineering with embeddings, that are learned from BiLSTMs \cite{cotterell2017low,suriyachaythai}. A CRF is placed at the end of the pipeline to handle the classification. A very different approach, hypothesizing that NE of the same type are embedded in the same region, showed that viewing the typing task as a clustering approach based on embeddings can outperform CRF architectures \cite{das2017named}. Phrases can easily be taken into account by this method. Finally \cite{mbouopda2020named} takes advantage of the low frequency of NEs in any document to derive a novel embedding for corpora aligned at the sentence-level. A traditional neural classifier is then fed with these embedding to perform classification. This contribution has the advantage of only requiring sentence alignment, and can be applied to other classification tasks where a few classes are more prevalent than many other.

\paragraph{NEL} Cross-lingual NEL (XEL) generally consists in two steps: candidate generation and candidate ranking. External sources of knowledge such as Wikipedia or Google Maps often help link or disambiguate entities \cite{gad2015named}. A feature-based neural model further improved by designing a new feature combination technique and \cite{zhou2019towards} proposed a new scheme to integrate two candidate generation methods (look-up- and neural-based), suggest a set of language-agnostic features and devise a non-linear combination method. This resulted in significant improvement in end-to-end NEL. \cite{zhou2020improving} replace the LSTM model by an n-gram bilingual model to solve the sub-optimal string modeling. Finally, \cite{fu2020design} relies on log queries, morphological normalization, transliteration and projection as a comprehensive improvement over even supervised methods.

\paragraph{Future work} As \cite{zamin2020projecting} reported, performance of NER systems is highly currently domain-specific ; some papers incentivize designing cross-domain techniques. The literature mentioned that conceiving a new metrics to compare cross-lingual embedding is promising avenue to XEL. Transferring from languages other than English would further allow to check the robustness of the currently developed methods.

\subsection{Morphology induction}

Morphology induction aims to align inflected word forms with their root form. This task unifies lemmatization and morphological paradigm completion, which is the reverse process, that is finding all inflected forms from a root word.

\begin{figure}
    \centering
    \includegraphics[width=\linewidth]{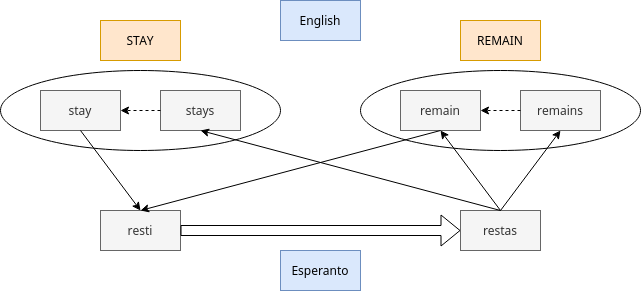}
    \caption{Morphology induction for Esperanto through English}
    \label{fig:morphology}
\end{figure}

The earliest method relied on word alignment between a source HRL and a target LRL and an existing morphology induction system in the source language \cite{yarowsky2001inducing}. Essentially, a word in the target language is mapped to its counterpart in the source language \emph{via} word-alignment, which redirects to its root form thanks to the morphology system. Word-alignment enables to map the source root form back into the target language to produce the desired output (Figure \ref{fig:morphology}).

A more advanced method makes the assumption that the word inflection occurs at the end of the word to derive a trie-based approach. If the inflected $i$ and root $r$ word share a longest common substring $s$ such that $i = s.\tilde{i}$ and $r = s.\tilde{r}$, then the probability that $i$ comes from $r$ is defined by

\[P(i|r) = \sum_{k=0}^{|r|}{\lambda_k P(\tilde{r} \to \tilde{i} | r_{<k})}\]

For example, $P(replies|reply) = \lambda_0 P(y \to ies) + \lambda_1 P(y \to ies|y) + \lambda_2 P(y \to ies|ly) + \lambda_3 P(y \to ies|ply) + \ldots$. Other methods to lemmatization involved learning edit-based strategies (copy, delete, replace) through LSTMs models like \cite{makarov2018neural}, the sequential nature of which strikingly outperforms character aligners.

Another more recent line of research has aimed to create morphological paradigms, that is the set of all rules that can be applied on a root form given its part-of-speech. The unsupervised process consists in three steps: candidate search, paradigm merging and generation. While a reductive approach relies on tries \cite{monson2007paramor}, a more robust method involves edit trees \cite{jin2020unsupervised} to model the transition between two words. Surrounding context is used to merge paradigms, and a transducer chooses from the different available shifts for a given slot to create the final output. For example in English, the third person of the present tense can be obtained by adding ``s'' or ``es''. The transducer chooses to apply ``es'' to the root ``miss'' rather than ``s''.

\paragraph{Future work}
This last work achieves strong performance in a very low-resource setting under the hypothesis that a morphological shift only appears in one paradigm. This is not always true (for example in English adding an ``s'' both enables to transition from singular to plural for nouns, and to create the third person verbal from from the infinitive). Future directions could include releasing this limiting assumption as well as making use of word embedding to merge paradigms and exploring other transduction models.

\section{Speech recognition}

The goal of large vocabulary continuous speech recognition (LVCSR) is to derive the meaning of a verbal message by identifying embedded phones. Applying LVSCR on LRLs primarily concerns how to exploit various language components such as phones and syllables while also employing transfer learning from high-resource secondary languages to a target LRL.

\begin{figure}
    \centering
    \includegraphics[width=0.8\linewidth]{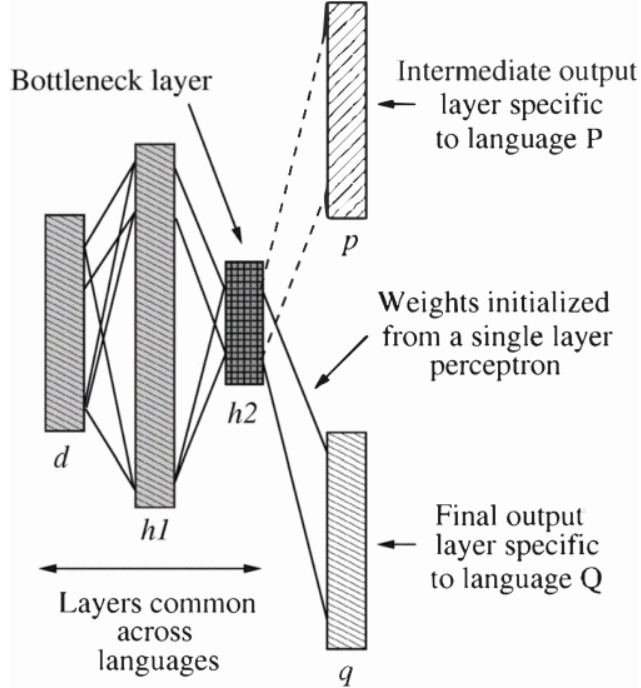}
    \caption{MLP model proposed by \cite{thomas2012multilingual} that replaces the last layer of a network trained on HRL 'P' with a single layer pre-trained on LRL 'Q'}
    \label{fig:mlp}
\end{figure}

\paragraph{Multilayer perceptrons}
A traditional approach for multilayer perceptron (MLP) networks in speech recognition is incorporating phonal training data from other HRLs and extracting tandem features for LRL classification. However, this originally required that data from all languages be transcribed using a common phoneset. The work in \cite{thomas2012multilingual} instead presented an MLP model that completed the same task without mapping each language to the same phoneset. As shown in Figure \ref{fig:mlp}, this is accomplished first by training a 4-layer MLP with layers $d$, $h_1$, $h_2$, and $p$ on HRLs. The last layer $p$ is then removed and replaced with a single-layer perceptron $q$ whose weights have been pre-trained using only limited data from the target LRL. The entire 4-layer model was lastly trained again using only target LRL data. This approach is roughly 30 percent more accurate than the baseline common-phoneset model.

\paragraph{Hidden Markov models}

Another popular speech recognition method is the hidden Markov model (HMM), which is a subclass of unsupervised dynamic Bayesian networks. An HMM in speech recognition essentially works by learning the probabilities for next sequential sounds given a particular sound in training speech. It is then made to confirm sequential sounds in speech on a phone, syllable, and word level. From an low-resource standpoint, HMM topology and the scope of speech modelling units are major points of consideration for ensuring success. For example, \cite{fantaye2019syllable} adopted a particular LRL's relatively few possible consonant-vowel syllables as modelling units and iterated through different HMM architectures. A deep-neural-network-based HMM using transfer learning from a language similar to the said LRL achieved greatest results. The same transfer learning paradigm for HMM-based speech recognition in LRLs also explored in \cite{adams2017cross}, which uses English as a source language for Turkish.

\paragraph{Future work}
One literature focus is performance of name and place entity detection in speech. One avenue of future work must address how to improve the detection accuracy of these speech components. Another concern is the lack of homogeneity among speech training data, which have various noise, speakers, accents, emotions, etc. that are difficult to standardize and create lots of variance. Further research should yield methods to filter and normalize training data.

\section{Embeddings}

Embeddings are the key to accurate and efficient NLP models. They represent a model's understanding of sentence structure and word meaning. However, training embeddings is a resource-heavy process and is not compatible with the data scarcity of LRLs. Hence, to avoid a random initialization of embeddings, different approaches are being used: multilingual representations, transfer learning for structurally similar languages, or even data augmentation.

\subsection{Data Augmentation}

\begin{figure}
    \centering
    \includegraphics[width=1\linewidth]{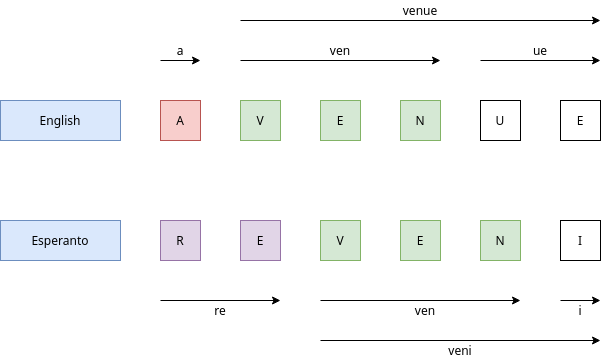}
    \caption{The radical ``ven'' is quite common in Esperanto, thus it could be integrated in the English-Esperanto segmentation, and benefit a few English word including ``venue'', ``avenue'', ``advent'' and others \cite{nguyen2017transfer}}
    \label{fig:BPE}
\end{figure}

Byte Pair Encoding segmentation, known as BPE segmentation, extracts additional data from the language corpus in order to generate the embeddings \cite{nguyen2017transfer}. The motivation behind BPE segmentation to take advantage of shared substrings: words from both dictionaries are broken down into subwords and then the most common subwords are kept as roots. (Figure \ref{fig:BPE}). This means that each word is therefore seen as a sequence of tokens (or subwords) which increases the amount of overlap between both vocabularies. This technique is particularly useful for LRLs as it improves the transferability of the embeddings. Indeed, having more tokens in common between the parent HRL and the target LRL means there are more overlapping words to align the embeddings during transfer.

\begin{figure}
    \centering
    \includegraphics[width=1\linewidth]{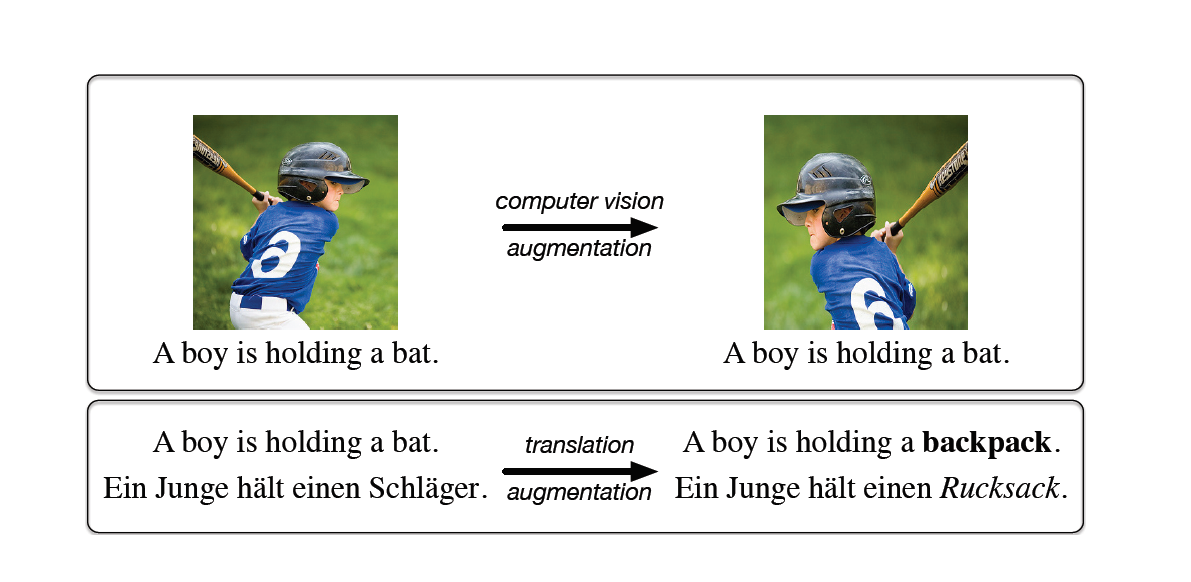}
    \caption{Comparison between Computer Vision and Low-Resource Languages data augmentation \cite{fadaee2017data}.}
    \label{fig:augmentation}
\end{figure}

To combat the training data scarcity, another approach is to transform a LRL into a HRL using data augmentation. A technique used in computer vision was adapted to NLP by \cite{fadaee2017data} (Figure \ref{fig:augmentation}).

The idea is to create additional sentences by replacing a single word in the sentences of the dataset. Still, the words replaced for data augmentation must fit the sentence structure properly (e.g. verb replacing another verb) in order for the generated sentence to be usable during training. However, even if the sentence structure is preserved, its meaning may be altered. This is problematic for multiple LRL domains and is rarely used outside of machine translation models. Indeed, for NMT, models are built to translate even meaningless sentences (e.g.: Boats are delicious!) as they only require proper word pairs translations and correct sentence structure.

\paragraph{Future work}
Although Data Augmentation techniques proved successful to combat several LRLs, current approaches struggle for extremely low-resource languages. In fact, to properly handle data augmentation, precise knowledge of sentence structure, grammar and words translation pairs is required in order to assert the correctness of the generated sentence.

\subsection{Multilingual Embeddings}

Multilingual embedding-based models have been introduced for zero-shot cross-lingual transfer. For instance, using a single Bidirectional Long Short Term Memory (BiLSTM) as the model encoder \cite{artetxe2019massively} allows to learn a classifier after training the multilingual embeddings. Therefore, the embeddings can be transferred to any of the languages the NMT is designed for without any structural modifications. 

Another multilingual embedding-based model architecture uses multilingual BERT to perform accurate zero-shot cross-lingual model transfer \cite{pires2019multilingual}. M-BERT proved effective at transferring knowledge across languages with different scripts (no lexical overlap), which proves that the model does capture multilingual representations into embeddings. These can later serve as a basis for multilingual tasks.

\section{Machine translation}

A Neural Machine Translation model, known as NMT model, aims to translate a sentence from a source language to a target language. NMT models are built using two separate but connected models: an encoder and a decoder. The encoder's goal is to break down the sentence logic and word pairs, and then store this information as embeddings. Afterwards, the decoder generates the translated sentence based on those embeddings.

This technology has proved to be highly accurate and reliable for translating between two HRLs. Yet, NMT models require large corpora of training data, based on translations or annotated data crafted by language experts. For common languages such as English, Spanish or French, the data used for training has already been processed in large quantities. However, when it comes to less common languages and dialects, there is little to none translated data available. That is why specific architectures and methods are required to handle LRL NMT. The state-of-the-art techniques and models are introduced in the following sections.

\subsection{Transfer Learning}

Transfer Learning \cite{zoph2016transfer} relies on the fact that NMT achieves great results for HRLs. Indeed, it uses a two-model architecture: a parent model which is a standard NMT model trained on a pair of HRLs (e.g. French-English or English-Spanish) and a child model which is the desired translation model between the source and target languages. More specifically, the parent model is trained using a standard corpus and then, the embeddings of the parent model are used to initialize those of the child model. This way, the embeddings of the child model are not randomly initialized and it can be trained more efficiently even with a small amount of data.

\paragraph{Future work} Transfer Learning works best when the parent and child languages have similar lexicon and grammar. Strategies to combat this issue still have to be developed, such as using an ordering model \cite{murthy2018addressing}. The goal of such a model is to reorder the sentences of the source language to match the structure of the target language. Having the same sentence structure allows for a more accurate transfer of embeddings as similar words will have similar positions in sentences.

\subsection{Prior Models}

A second approach to cope with LRLs consists in introducing a prior model during training. Indeed, the Zero-Shot Translation technique \cite{kumar2019augmented} allows to train the encoder model using multiple languages or dialects simultaneously. This model is able to capitalize on the already learned languages pairs to translate between unseen language pairs. For instance, if the model has been trained for English - French and French - Chinese pairs then it can already handle English - Chinese translation. Thus avoiding the rebuilding of the NMT system for every new language pair.

Further works have replaced the Zero-Shot Translation model by a Language Models (LM) \cite{baziotis2020language}. The LM adds a regularization term that pushes likelier output distributions to the NMT. This means that, out of the top predictions output by the NMT, only the top word amongst the ones which are validated by the LM will be selected. Hence, this can be seen as a knowledge distillation technique where the LM is teaching the NMT about the target language. 

\begin{figure}
    \centering
    \includegraphics[width=1\linewidth]{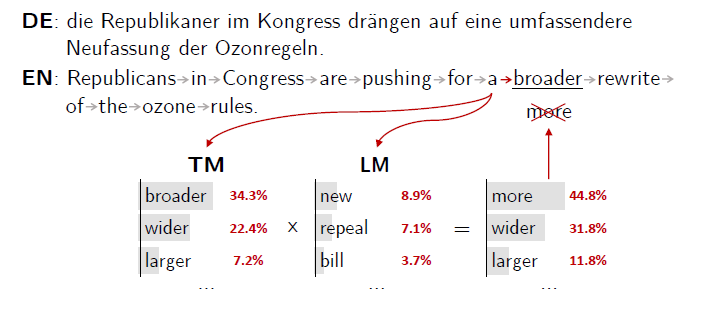}
    \caption{Example of an incorrect prediction when the LM and NMT model disagree \cite{baziotis2020language}.}
    \label{fig:prediction}
\end{figure}

\paragraph{Future work}
Language Model-based NMT, however, is still exposed to incorrect predictions in some cases where the LM and the NMT model disagree on a prediction, even when the NMT model predicted correctly (Figure \ref{fig:prediction}).

\subsection{Multilingual Learning}

Multilingual Learning extends the transfer learning techniques in multilingual environments. The idea is to build an NMT model using a universal shared lexicon and a shared sentence-level representation, which is trained using multiple LRLs. One of the first approaches \cite{gu2018universal} consists in using two additional components compared to the traditional transfer learning techniques. The first one is a universal lexical representation to design the underlying word representation for the shared embeddings. The second is a mixture of language experts to deal with the sentence-level sharing. This system is used during the encoding and works as a universal sentence encoder. It allows to transfer the learned embeddings for a given language pair into the universal lexical representation.

A Model-agnostic meta-learning (MAML) algorithm applied to low-resource machine translation allows to view language pairs as separate tasks \cite{gu2018meta}. This technique results in faster and more accurate training of the vocabulary. Although, as MAML was initially designed for deep learning purposes, it cannot handle tasks with mismatched input and output. That is why it is only applied to Universal NMT models.

\paragraph{Language Graph}

A rather structurally different approach introduces the concept of Language Graph for Multilingual Learning \cite{he2019language}. Vertices represent the various LRLs and edges represent the translation word pairs. Moreover, for each edge, a weight score is assigned to denote the accuracy of the translation pair. Then, a distillation algorithm is used to maximize the total weight (accuracy) of the model, using forward and backward knowledge distillation to boost accuracy. The main advantage of such a graph is that there exist multiple translation paths from one word to another. (e.g. English to Spanish and English to French to Spanish). For LRLs, the direct translation path generally has low accuracy due to the lack of available parallel data. However, there are some languages for which the translation pair will have high accuracy and this resulting path will be put to use. 

\section{Classification}

\begin{figure}
    \centering
    \includegraphics[width=1\linewidth]{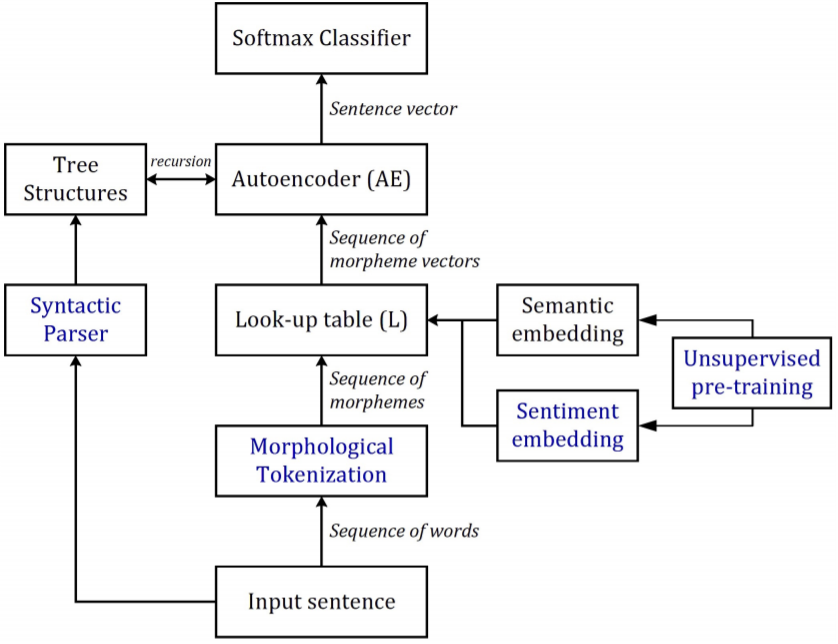}
    \caption{Aroma framework proposed by \cite{al2017aroma}, with augmented portions in blue}
    \label{fig:aroma}
\end{figure}

\paragraph{Sentiment analysis} Advanced opinion mining models designed for the English language do not work well on LRLs that have vastly different grammar, unstructured format, and little applied NLP research or resources. From a machine learning standpoint, one alternative is adopting lighter, less resource-dependent baseline models that are still successful with the English language and modifying them to maximize success with a particular LRL. For example, \cite{al2017aroma} adopted a recursive auto encoder (RAE) baseline model for Arabic and augmented it with morphological tokenization, a sentiment-extracting neural network, an unsupervised pre-training block to improve sentiment embedding initialization, and a phrase structure parser to generate parse trees (Figure \ref{fig:aroma}).

An alternative solution is the lexicon-based approach, which manually classifies words in a sentence based on a list of sentimentally-classed words. The lexicon-based method, various machine learning methods such as K Nearest Neighbor, Naive Bayes, and Decision Tree, as well as hybrid architectures between the two are applied on the English and Urdu languages in \cite{azamsentiment}.

\paragraph{Data expansion} One way to lift LRLs out of the ``low-resource'' category is obviously to increase the quality and quantity of available supervised classification data. One such approach is to expand existing data through adversarial distortion of text attributes, or by extracting more features using a transfer learning technique that surmount pre-trained layers from recognition systems for multiple common languages with new trainable layers \cite{qi2019study}. A more direct approach is to manually compile a corpus dataset that includes multi-label text classifications and pre-training language models as a platform for further NLP work for respective LRLs as done in \cite{cruz2020establishing} for the Filipino language.
 
\paragraph{Miscellaneous} There are many more LRL classification sub-topics. One such avenue is text readability classification which, for example, can automate quality analysis and pinpoint areas requiring edit in LRL textbooks using lexical, entropy-based, and divergence-based features \cite{islam2012text}.
    
Pattern recognition is another problem for LRLs due to the lack of NLP study on coping with less common textual attributes. The work in \cite{abliz2020research} has addressed impediments such as vowel weakening and suffix-based morphological changes for the Uyghur language through a proposed algorithm that performs pattern matching using syllable features.
 
\paragraph{Future work} Although the above literature had successful approaches for addressing the LRL classification problem in their respective areas, the overall research direction is still rooted in the fundamental understudy and lack of experience with LRLs. Future classification work seems to be be focused on two general categories. One line will be further study of LRL morphological traits and grammar patterns to increase model performance. Another branch of research will focus on applying existing classification models to use as a benchmark to improve off of for more novel methods. 
 
\section{Discussion}

On top of the low-level examination we conducted for various NLP tasks, we would like to summarize two desiderata recurrently noted the literature: collecting new datasets for more diverse languages and devising a closeness index for languages.

\paragraph{Datasets diversity}
A few papers collected new datasets in innovative ways, that we believe should be further put forward. Extracting news headlines or comments from social media \cite{marivate2020investigating}, relying on mobile applications to gather audio extracts and annotations \cite{godard2017very}, as well as relying on governmental sources, constitute new alleys to dataset creation. Finally, we would like to mention the Tatoeba \footnote{\url{www.tatoeba.org}} project, that is a collaborative, open and free collection of aligned sentences and translations in more than 350 languages. 

\paragraph{Closeness index for languages} Throughout our review, we encountered a fair amount of papers that reported the difficulty to select language pairs that allow for smooth transfer or alignment. As quoted by \cite{nasution2017generalized}, the Automated Similarity Judgment Program (ASJP) \cite{asjp2020} \footnote{\url{www.asjp.clld.org/}} has collected a word list based on the Swadesh list for more than 9500 languages and dialects. This allows for a morphological and lexical comparison, yet cannot help match grammatical differences across languages. That is why we advocate the design of a task-specific linguistic distance that would model both the morphological and grammatical aspects of a language, and, given a target language, would guide the choice of the optimal language to transfer from.

\section{Conclusion}

A review of over 60 LRL-related papers has yielded a general idea of this field's most recent work. In an area that concerns a fundamental lack of data, a primary trend is expanding LRLs datasets while also applying augmentation methods and transfer learning techniques from other languages in a manner that copes with their differences. Future work will highly involve improving the quality of LRLs data, taking advantage of linguistic patterns/similarities, designing more robust learning models, and increasing the reliability of evaluation methods.

\bibliographystyle{acl_natbib}
\bibliography{mybib}

\end{document}